\documentclass[sigconf]{acmart}
\usepackage{adjustbox}
\usepackage{xcolor}

\AtBeginDocument{%
  \providecommand\BibTeX{{%
    \normalfont B\kern-0.5em{\scshape i\kern-0.25em b}\kern-0.8em\TeX}}}

\setcopyright{acmcopyright}
\copyrightyear{2018}
\acmYear{2018}
\acmDOI{XXXXXXX.XXXXXXX}

\acmConference[Conference acronym 'XX]{Make sure to enter the correct
  conference title from your rights confirmation email}{June 03--05,
  2018}{Woodstock, NY}
\acmPrice{15.00}
\acmISBN{978-1-4503-XXXX-X/18/06}

\begin{document}

\title{Graph Topology Information Enhanced Heterogeneous Graph Representation Learning}

\author{He Zhao}
\affiliation{%
  \institution{Nanyang Technological University}
  \institution{Joint NTU-UBC Research Centre of Excellence in Active Living for the Elderly}
  \streetaddress{50 Nanyang Avenue}
  \country{Singapore}
  \postcode{639798}
}
\email{ZHAO0378@e.ntu.edu.sg}

\author{Zhiwei Zeng}
\affiliation{%
  \institution{Nanyang Technological University}
  \institution{Joint NTU-UBC Research Centre of Excellence in Active Living for the Elderly}
  \streetaddress{50 Nanyang Avenue}
  \country{Singapore}
  \postcode{639798}
}
\email{zhiwei.zeng@ntu.edu.sg}

\author{Yongwei Wang}
\affiliation{%
  \institution{Nanyang Technological University}
  \streetaddress{50 Nanyang Avenue}
  \country{Singapore}
  \postcode{639798}
}
\email{yongweiw@outlook.com}

\author{Chunyan Miao}
\affiliation{%
  \institution{Nanyang Technological University}
  \institution{Joint NTU-UBC Research Centre of Excellence in Active Living for the Elderly}
  \streetaddress{50 Nanyang Avenue}
  \country{Singapore}
  \postcode{639798}
}
\email{ASCYMiao@ntu.edu.sg}


\begin{abstract}
Recent studies indicate that the performance of GNNs largely depends on the quality of the input graph structure. Real-world heterogeneous graphs are inherently noisy and usually not in the optimal graph structures for downstream tasks, which often adversely affects the performance of GRL models in downstream tasks. Although Graph Structure Learning (GSL) methods have been proposed to learn graph structures and downstream tasks simultaneously, existing methods are predominantly designed for homogeneous graphs, while GSL for heterogeneous graphs remains largely unexplored. Two challenges arise in this context. Firstly, the quality of the input graph structure has a more profound impact on GNN-based heterogeneous GRL models compared to their homogeneous counterparts. Secondly, most existing homogenous GRL models encounter memory consumption issues when applied directly to heterogeneous graphs. 
In this paper, we propose a novel Graph \underline{\textbf{To}}pology learning Enhanced Heterogeneous \underline{\textbf{G}}raph \underline{\textbf{R}}epresentation \underline{\textbf{L}}earning framework (ToGRL). 
ToGRL learns high-quality graph structures and representations for downstream tasks by incorporating task-relevant latent topology information. Specifically, a novel GSL module is first proposed to extract downstream task-related topology information from a raw graph structure and project it into topology embeddings. These embeddings are utilized to construct a new graph with smooth graph signals. This two-stage approach to GSL separates the optimization of the adjacency matrix from node representation learning to reduce memory consumption. Following this, a representation learning module takes the new graph as input to learn embeddings for downstream tasks. ToGRL also leverages prompt tuning to better utilize the knowledge embedded in learned representations, thus enhancing adaptability to downstream tasks. Extensive experiments on five real-world datasets show that our ToGRL outperforms state-of-the-art methods by a large margin. In addition, ToGRL achieves good performance in terms of memory consumption, addressing the out-of-memory issues prevalent in existing GSL models.

\end{abstract}





\maketitle

\section{Introduction}
Graphs are widely used data structures for modeling complex real-world interaction systems, such as social networks, recommender systems, and knowledge graphs.
Graph Representation Learning (GRL) aims to embed graph information into low-dimension representations while preserving important graph topology and node attributes. 
Recently, Heterogeneous GRL (HGRL) models have been developed to incorporate heterogeneous structural and attribute information, achieving remarkable performances on various downstream tasks, such as node classification \cite{yun2019graph,hu2020heterogeneous}, link prediction \cite{fu2020magnn,wang2022collaborative}, and recommendation \cite{lv2021we}. 

Given the superiority of Graph Neural Networks (GNNs) in learning node representations, existing GRL methods often adopt GNNs as learning backbones \cite{kipf2016semi, velivckovic2017graph, xu2018powerful}. Studies have shown that GNNs are highly sensitive to the quality of input graph structures~\cite{franceschi2019learning, wang2023revisiting}, and hence, noisy or suboptimal input graphs often lead to unsatisfactory graph representations. 
Most GNNs are designed with the implied hypothesis that the input graphs are reliable and contain adequate information related to downstream tasks~\cite{franceschi2019learning,zhao2021heterogeneous}. 
However, in real-world applications, the quality of raw input graphs is usually poor. For instance, in the e-commerce recommendation scenario, a user may mistakenly click on items of no interest, resulting in a noisy user-item interaction graph. Additionally, raw graphs are often not optimized for downstream tasks, as they are typically extracted using manually defined rules while overlooking the requirements of downstream tasks. This process may also introduce redundant information into graphs \cite{zhao2021heterogeneous}. 

The quality of the input graph structure has an even more profound impact on GNN-based HGRL methods than on homogeneous GRL methods. Existing HGRL models typically learn embeddings from different meta-paths separately to capture various semantic meanings~\cite{dong2017metapath2vec} and then fuse them into final embeddings \cite{wang2019heterogeneous,fu2020magnn}. Consequently, the use of the meta-paths mechanism, in a sense, proliferates the noises in the original input graphs, introducing additional noises to HGRL models.

On the one hand, a single noisy node can generate multiple pairs of meta-path-based neighbors, which are then used as inputs to HGRL models~\cite{wang2019heterogeneous,fu2020magnn, wang2022collaborative}. This ``hub'' effect~\cite{zhang2022robust} enlarges the effect of the single noisy node and degrades model performance. On the other hand, meta-path-induced homogeneous graphs often contain incomplete information. These problems will be further discussed in Section~\ref{sec:mining}.
As a result, HGRL models are more susceptible to noise in original input graphs than homogeneous GRL models. 

Therefore, to improve the quality of representations learned from heterogeneous graphs and enhance downstream task performance, it is essential to learn high-quality input graphs that contain less noise and are optimized for downstream tasks.
This need has motivated the line of research in Graph Structure Learning (GSL), which aims to learn an optimized graph structure and its corresponding representations simultaneously. GSL has been applied to homogeneous graphs and achieved very promising results~\cite{chen2020iterative,fatemi2021slaps}. However, these efforts on homogeneous graphs can not be easily transferred to HGRL, as they suffer from severe memory consumption problems when directly applied to heterogeneous graphs.

In this work, we propose a novel Graph \underline{\textbf{To}}pology Learning Enhanced Heterogeneous \underline{\textbf{G}}raph \underline{\textbf{R}}epresentation \underline{\textbf{L}}earning framework (ToGRL). 
To learn high-quality graph structures and representations optimized for downstream tasks, ToGRL utilizes downstream task-relevant latent topology information to guide the representation learning process. This approach allows for capturing more underlying mechanisms of downstream tasks in the learned representations. Specifically, our ToGRL employs a two-stage approach that separates the optimization of the adjacency matrix from node representation learning. First, our ToGRL utilizes a graph structure learning module to extract downstream task-relevant latent topology information into topology embeddings. Inspired by the graph theory in Graph Signal Processing (GSP), the learned topology embeddings are utilized to construct a new sparse graph in which the graph signals are smooth \cite{dong2019learning,humbert2021learning}. Then, a representation learning module takes the input as the newly constructed graphs to learn the final embeddings. This two-stage approach significantly improves memory efficiency by avoiding the need to model the entire adjacency matrix of the new graph as parameters that are jointly trained via GNNs based on the labels of downstream tasks.

To further enhance the performance on downstream tasks, ToGRL also innovatively introduces prompt-tuning into the graph domain, which better utilizes the knowledge embedded in learned representations for downstream tasks. During the downstream evaluation stage, we find that the number of labels is often limited in downstream tasks, which means a simple decoder may not be fully trained and fails to effectively utilize the knowledge learned during representation learning. In other words, there is a gap between graph representation learning and downstream tasks in the graph domain. Recently, Prompt Tuning \cite{gu2021ppt,liu2021pre} has been proposed as a new paradigm to narrow the gap between pre-training and fine-tuning in Natural Language Processing (NLP). By designing proper prompt templates, model performance can be significantly improved in limited label settings. Inspired by the success of prompt tuning in NLP, we propose to leverage a prompt-tuning inspired downstream task decoder for the heterogeneous graphs representation learning task that can further improve downstream task performance through more effective utilization of the learned representations. A task-specific prompt embedding is incorporated and jointly learned with the decoder. 

Extensive experiments are performed on five real-world datasets on two downstream tasks and show that our proposed method outperforms existing baselines. Ablation studies reveal that the new graph can obtain all the meta-path information and can better facilitate learning embeddings for downstream tasks. The prompt-enhanced decoder can efficiently extract knowledge contained in the embeddings. Besides, analysis results show that our framework can also significantly reduce memory consumption compared with GSL methods to the level of traditional HGSL methods.

Our major contributions can be summarized as follows:
\begin{itemize}
    \item We propose a novel heterogeneous graph representation learning framework enhanced by graph topology learning that can be trained efficiently in a two-stage learning process.
    
    \item We design a novel graph structure learning module that can effectively mine latent graph topology information together with downstream tasks. A new graph containing both topology and downstream task information is learned to facilitate learning downstream task embeddings. 
    
    \item To bridge the gap between embedding learning and downstream tasks,
    we propose to leverage prompt tuning for heterogeneous graphs on downstream tasks to help the decoder effectively utilize the knowledge of embeddings. 
    
    \item We perform extensive experiments on five real-world datasets on node classification and link prediction tasks. Experimental results and ablation studies show the effectiveness of the proposed HGRL framework.
\end{itemize}

\section{Related work}
In this section, we present a brief review of existing studies related to our work. In Section \ref{sec:GRL}, we overview existing typical heterogeneous graph representation learning methods. In Section \ref{sec:GSL}, we summarize related graph structure learning methods and then introduce prompt tuning methods in Section \ref{sec:Prompt}.

\subsection{Graph Representation Learning} \label{sec:GRL}

 Graph representation learning aims to learn a low-dimensional embedding for each node in a graph that can be used in downstream tasks. Graph Neural Networks (GNNs) have achieved promising results following a message-passing scheme, i.e., GNNs aggregate and transform neighbors information of each node \cite{kipf2016semi, velivckovic2017graph, xu2018powerful}. However, these models are designed for homogeneous graphs, which ignore the rich heterogeneity information in heterogeneous graphs. Recently, many heterogeneous graph representation learning methods have been proposed to address this problem. Metapath2Vec \cite{dong2017metapath2vec} utilizes random walks sampled according to meta-paths to model the heterogeneous information following a homogeneous graph embedding learning scheme \cite{grover2016node2vec}. To utilize the powerful learning ability of GNNs, \textbf{Heterogeneous Graph Neural Networks} (HGNNs) are developed. HAN \cite{wang2019heterogeneous} and HGT \cite{hu2020heterogeneous} are representative HGNNs that learn embeddings from meta-paths induced homogeneous graphs. MAGNN \cite{fu2020magnn} follows the message-passing scheme of HGNNs and extends HAN by considering not only the meta-path-based neighbors but also neighbors along each meta-path instance. SimpleHGN \cite{lv2021we} adopts a simpler scheme that abandons meta-paths and extends on GAT by incorporating edge embeddings of different relation types. The success of SimpleHGN also doubts the efficacy of the usage of meta-paths. 
 Recent efforts have been made to develop \textbf{Heterogeneous Graph Representation Learning} (HGRL) methods which are unsupervised using mutual information. HeCo \cite{wang2021self} proposes to use a cross-view contrastive learning method on network schema views and meta-path views of a same heterogeneous graph. HGIB \cite{yang2021heterogeneous} uses Information Bottleneck (IB) theory to maximize mutual information related to labels while minimizing the specific information between different meta-path induced homogeneous graphs. CKD \cite{wang2022collaborative} adopts knowledge distillation to jointly learn embeddings among different meta-paths and within each meta-path. 
 
 Despite the success of HGNN and HGRL models, they are developed based on an assumption that the input graphs are reliable. However, real-world heterogeneous graphs are usually noisy and may not be the optimal graph structure for downstream tasks, which often limits the performance of HGRL models.
 
\subsection{Graph Structure Learning} \label{sec:GSL}

Real-world graphs are usually noisy and there usually exists a gap between the extraction of graphs and the requirement of downstream tasks. Besides, some datasets do not have a readily available graph structure. Recently, Graph Structure Learning (GSL) has been proposed to address these challenges. To infer a new graph structure, one typical type of approach is to construct a sparse top-k Nearest Neighbor (kNN) graph based on a certain similarity metric. In \cite{gidaris2019generating}, a fixed kNN graph is constructed by calculating the cosine similarity of node features. CoGSL \cite{liu2022compact} first fuses different basic views such as adjacency matrix and diffusion matrix, and learns a minimal sufficient graph structure based on mutual information compression. HGSL \cite{zhao2021heterogeneous} focuses on heterogeneous graph structure learning by constructing multiple similarity graphs and semantic graphs and fusing them via attention weights together with the original graph structure. Another popular type of method tries to directly optimize the adjacency matrix which can be jointly trained with downstream tasks. An adjacency matrix is first initialized as a full parameter (or generated by an MLP), then a sparsification operation is applied to obtain the adjacency matrix used as input for GNNs. SLAPS \cite{fatemi2021slaps} infers the new graph by jointly training with a downstream node classification task and a node feature reconstruction task. IDGL \cite{chen2020iterative} learns the graph structure and node embeddings iteratively. SUBLIME \cite{liu2022towards} adopts contrastive learning on a learner view and an anchor view in an unsupervised manner. 

A main drawback of the first type of method is that the learned graph depends on the manually defined fixed graphs (e.g., similarity graphs) which cannot automatically generate new edges that do not exist in the input graphs. The second type of approach usually suffers from the memory consumption problem as they directly optimize on the adjacency matrix that is often of huge size in real applications. 

\subsection{Prompt Tuning} \label{sec:Prompt}
Pre-training Language Models (PLMs) \cite{devlin2018bert,liu2019roberta} have achieved promising performances in natural language processing. PLMs follow a `pre-training and fine-tuning' paradigm, in which the models are first pre-trained on a large corpus as language models and then fine-tuned with different downstream tasks. Objective engineering \cite{liu2021pre} is often applied to make LMs adapt to downstream tasks. These methods have a drawback that the pre-training knowledge is not fully utilized and the performance is not satisfying especially when sufficient training data is not available. Recently, prompt tuning methods \cite{petroni2019language,schick2020s,gao2020making,liu2023pre} are proposed to effectively utilize the knowledge learned during the pre-training stage and close the gap between pre-training and fine-tuning. By designing proper prompts, the pre-trained language model can be directly used to give the desired output and can perform well using limited labels. Early works \cite{davison2019commonsense,jiang2020can} apply a manually defined textual template 
(a.k.a discrete or hard prompts), e.g., `I felt so \_ ' and let the language model fill the slot with an emotional world. One limitation is that the hard prompt needs to be specially designed for different tasks. Recently, soft prompts \cite{li2021prefix,zhong2021factual} are proposed to automatically learn a continuous template that can let the language model fit for downstream tasks. There also exist works \cite{liu2021gpt,han2021ptr} that use a hybrid prompt tuning method that inserts trainable parameters into hard templates instead of using a purely learnable soft template. 

In this work, we propose to adopt a heterogeneous graph prompt tuning method to narrow the gap between heterogeneous graph representation learning and downstream tasks.

\section{Preliminaries}
In this section, we give formal definitions of concepts and essential notations related to this work.

\textbf{Definition 1. Heterogeneous Graph}. A heterogeneous graph is defined as $\mathcal{G}=\{\mathcal{V}, \mathcal{E}, X,\phi,\psi\}$, where $\mathcal{V}$ is a node set, $\mathcal{E}$ denotes an edge set and $X \in \mathbb{R}^{\left|\mathcal{V}\right|\times d}$ represents a node feature matrix where $d$ denote the dimension of a node feature. $\phi$ and $\psi$ denote a node type mapping function $\phi: \mathcal{V} \rightarrow \mathcal{T}$ and an edge type mapping function $\psi: \mathcal{E} \rightarrow \mathcal{R}$, where $\mathcal{T}$ and $\mathcal{R}$ are predefined node type set and edge type set, respectively. A heterogeneous graph satisfies a restriction that $\left|\mathcal{T}\right|+\left|\mathcal{R}\right|>2$; otherwise, it degrades to a homogeneous graph. 


\textbf{Definition 2. Meta-path}. Meta-paths have been widely used to extract heterogeneous information with specific semantic meanings. A meta-path is manually defined in the form of $T_1\stackrel{R_1}{\longrightarrow}T_2\stackrel{R_2}{\longrightarrow} \cdots \stackrel{R_k}{\longrightarrow}T_{k+1}$, usually represented as $T_1 \circ T_2 \circ \cdots \circ T_{l+1}$, where $\circ$ denotes a composite operation and $T_i \in \mathcal{T}, R_j \in \mathcal{R}$. Usually, the head node type and the tail node type are the same in a meta-path. 
A meta-path instance is a sampled node sequence according to a meta-path.
Given a meta-path $P$, a meta-path induced homogeneous graph $\mathcal{G}_P$ indicates that, for each edge in $\mathcal{G}_P$, there exists at least one meta-path instance between the head node and tail node. 

\textbf{Definition 3. Graph Neural Networks}. GNNs aim to project node information into low-dimensional vectors $h_i \in \mathbb{R}^d$ where $d$ is the embedding dimension. GNNs follow a message-passing scheme that information on each node is aggregated from its neighboring nodes to update its own embedding. Graph Convolution Network (GCN) and Graph Attention Network (GAT) \cite{kipf2016semi} are representative GNNs, which are briefly introduced as follows. 
\begin{figure*}[htbp]
    \centering
    \includegraphics[width=0.9\linewidth]{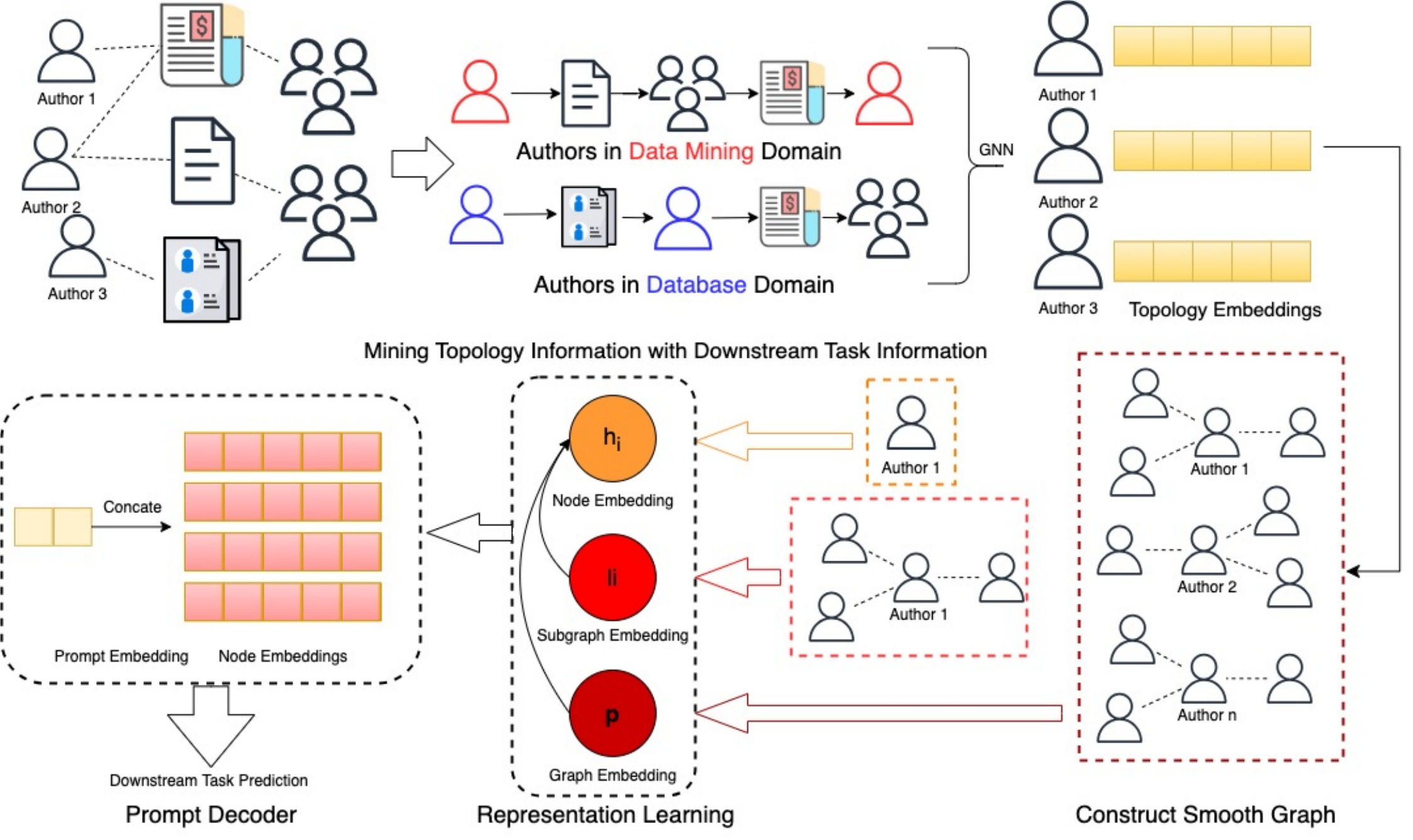}
    \caption{An overview of our ToGRL framework. First, a graph structure learning module is used to mine latent graph topology together with downstream task information. The learned topology embeddings are utilized to construct a new graph. Then, a representation learning module is used to learn final node representations. Finally, a prompt tuning decoder is used to evaluate embeddings on downstream tasks.}
    \label{fig:overview}
\end{figure*}

\textbf{Graph Convolution Network} (GCN). In a GCN, the $l$-th layer can be mathematically expressed as:
\begin{equation}
    \textbf{H}^{(l)} = \sigma(\hat{\textbf{{A}}}\textbf{H}^{(l-1)}\textbf{W}^{(l)})
\end{equation}
\begin{equation}
    \hat{\textbf{A}} = {\textbf{D}}^{-\frac{1}{2}}(\textbf{A}+\textbf{I})\textbf{D}^{-\frac{1}{2}}
\end{equation}
where $\hat{{A}}$ is a normalized adjacency matrix, $H^{(l)}$ is the updated node representation  and $W^{(l)}$ is the weight matrix in the $l$-th layer. 

\textbf{Graph Attention Network} (GAT). GAT extends GCN by replacing the average operation in each layer by attention weights over each edge $e_{ij}$:
\begin{equation}
    \alpha_{ij} = \frac{\mathrm{exp}(\mathrm{LeakyReLU}(a^T[\textbf{W}h_i||\textbf{W}h_j]))}
    {\sum_{k\in \mathcal{N}_i}\mathrm{exp}(\mathrm{LeakyReLU}(a^T[\textbf{W}h_i||\textbf{W}h_k]))}
\end{equation}
where $W$ and $a$ are learnable parameters, $|\cdot|$ denotes a concatenation operation, and $\mathcal{N}_i$ is the neighbor node set of node $i$. Note that homogeneous graph neural networks can also handle heterogeneous graphs by treating node type and edge type to be the same. 

\section{Methodology}
In this section, we propose a novel \underline{\textbf{To}}pology learning Enhanced heterogeneous \underline{\textbf{G}}raph \underline{\textbf{R}}epresentation \underline{\textbf{L}}earning framework (ToGRL). The ToGRL method has three major steps: 1) mining latent graph topology information and constructing a new graph; 2) learning node representations based on the new graph; and 3) evaluating embeddings using a prompt tuning enhanced decoder. We describe the details of each step in the following sections. An overview of the whole framework is given in Fig.~\ref{fig:overview}. 

\subsection{Mining Latent Graph Topology Information}
\label{sec:mining}
GNNs have achieved promising results in recent years and have shown their powerful expressive capability in learning node representations by aggregating information from neighborhood nodes. Thus, most existing HGRL models adopt GNNs as learning backbones to learn heterogeneous graph representations. However, simply applying GNNs on the raw heterogeneous graph is not the optimal choice for two reasons:

\begin{figure}[tb]
    \centering
    \includegraphics[width=1.0\linewidth]{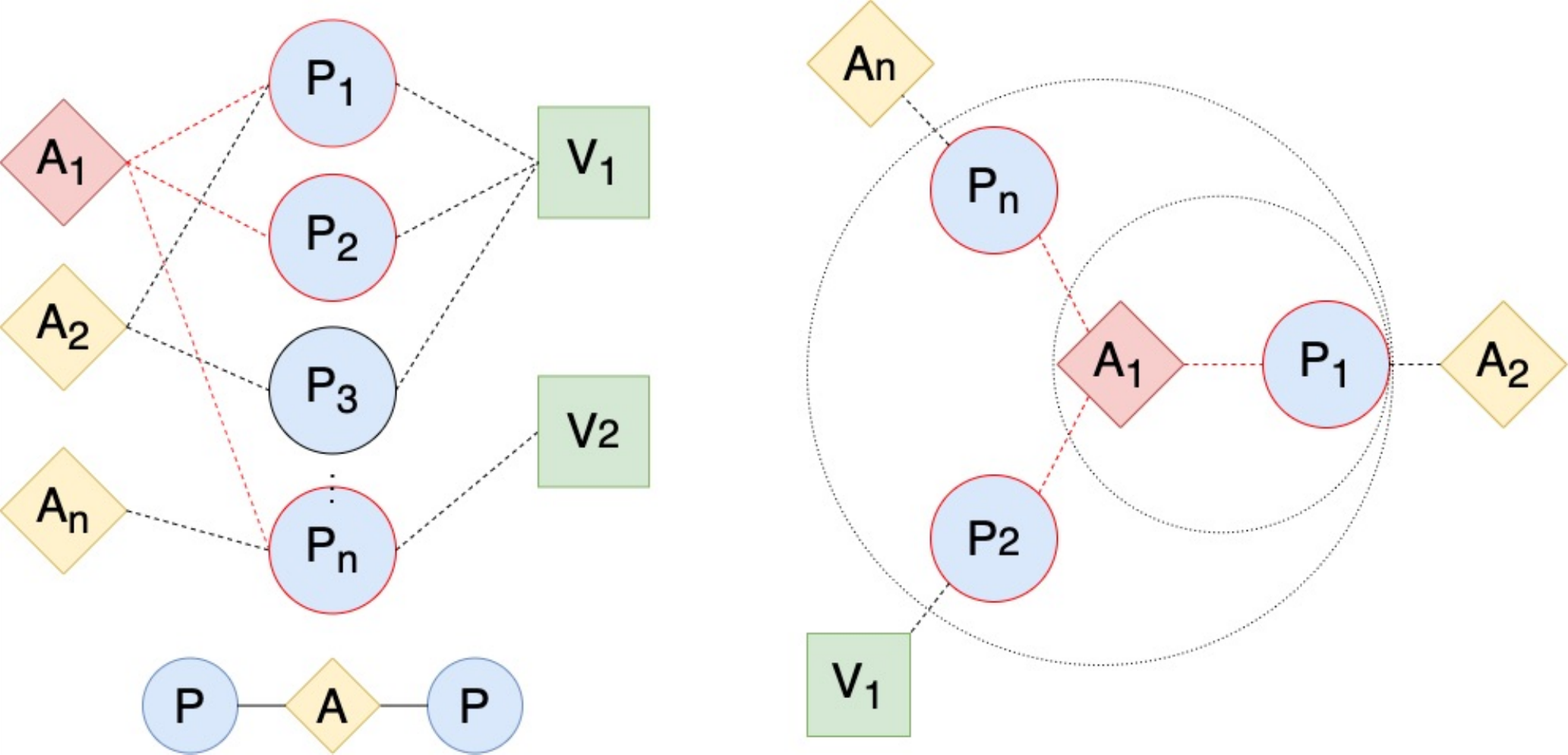}
    \caption{Noise effect is enlarged by changing from two-hop relationships (in homogeneous models on the right) to one-hop relationships(in heterogeneous models on the left).}
    \label{fig:hub}
\end{figure}

\textbf{1. Noisy meta-path-based neighbors.} As proposed in Zhang's work\cite{zhang2022robust}, the ``hub'' effect will enlarge the effect of one noise node by generating multiple meta-path-based neighbor pairs.  For example, in the Fig \ref{fig:hub}, one noise node $A_1$ can 
generate meta-path-based neighbor pairs, e.g. $(P_1, P_3), (P_2, P_3)$, making these nodes have direct relationships within each other. However, in homogeneous GNNs, the relationship is two-hop. To this end, the noise effect is enlarged in heterogeneous models.



\textbf{2. Incomplete information of meta-paths induced homogeneous graphs}. Most HGRL methods \cite{wang2022collaborative,hu2020heterogeneous,wang2019heterogeneous} apply GNNs on several meta-paths induced homogeneous graphs. Although such homogeneous graphs well represent each meta-path information, specific information (e.g., \textit{Author A writes Paper P in Venue V}) is ignored in such a homogeneous graph. In this case, the model only knows the co-venue relationship among authors while it has no idea how a pair of authors have such relations. Specifically, there are two cases that two authors have a co-author relationship: one is that the authors have a direct co-author relationship, and the other is that they contribute to different papers but their papers are in the same venue. Apparently, two authors in the first case have more chances to be in the same research field than that in the second case. Thus, models learned from these graphs will lead to a degradation in performance.

To learn graphs with less noise and fully extract semantic information, we propose a \textbf{Graph Structure Learning Module} based on a novel labeled random walk strategy by exploring the latent topology information of the original graph together with downstream tasks information. Different from existing GSL methods \cite{fatemi2021slaps,liu2022compact} which jointly optimize the graph structure and downstream task,  our ToGRL operates in a two-stage manner which is memory effecient by firstly learning one new graph and then node embeddings are learned in the following module based on the fixed new graph. 
Furthermore, in our framework, we aim to learn a new unified graph that is amenable for downstream tasks, thus we sample paths using Node2Vec\cite{grover2016node2vec} instead of Metapath2Vec\cite{dong2017metapath2vec}, which sample paths based on one given meta-path (there can be multiple meta-paths for one dataset).

We first extract latent graph topology information and project it into \textit{\textbf{Topology Embeddings}} based on random walk.
Our intuition is that a path that is fully walked on the graph can contain the meta-path information and preserve the above missing information. Given a heterogeneous graph $\mathcal{G}=\{\mathcal{V}, \mathcal{E}, X,\phi,\psi\}$, we apply the walking strategy used in Node2Vec \cite{grover2016node2vec}, in which the walks are generated by:
\begin{equation}
    P(c_i=a|c_{i-1}=b) = \left\{ \begin{array}{rcl} \frac{\pi_{ab}}{Z} & \mathrm{if} \;(a,b)\in E \\
    0 & \mathrm{otherwise}
    \end{array}\right.
\end{equation}
\begin{equation}
    \pi_{ab} = \alpha_{pq}(t, a)=
    \left\{ \begin{array}{rcl} 
    \frac{1}{p} & \mathrm{if} \;d_{ta}=0 \\
    1 & \mathrm{if} \;d_{ta}=1 \\
    \frac{1}{q} & \mathrm{if} \;d_{ta}=2\\
    \end{array}\right.
\end{equation}
where $\pi_{ab}$ is the transition probability, $Z$ is a normalized constant, $p, q$ are hyper-parameters and $d_{ta}$ is the shortest distance between node $t$ and $a$. Assuming that the walk has just traversed edge $(t,a)$, $p$ controls the likelihood of immediately revisiting a node in the walk, and $q$ controls the walking preference between Breadth-first Sampling(BFS) and Depth-first Sampling (DFS). 

Given the sampled random walks, the training objective is a skip-gram loss which maximizes the log probability that observing node $v$'s neighborhood $N_v$ based on its feature. In practice, we apply negative sampling and the objective loss function is defined as: 
\begin{equation}
    \min_{f} \sum_{v\in V}\left(\sum_{u\in N_{v}^+}-\mathrm{log}(f(v)\cdot f(u))
                                +\sum_{k\in N_{v}^-}-\mathrm{log}(1-f(v)\cdot f(k))
                                \right)
\end{equation}
where $N_v^+$ is the positive neighbors based on the given walks and $N_v^-$ is the negative set, $f(v)$ is the learned topology embedding through a graph neural network. In this work, we mainly focus on applying GCN \cite{kipf2016semi} and GAT \cite{velivckovic2017graph} as typical learning backbones. Yet, our choice can be readily extended to other types of GNNs. Besides, the random walk strategy can be extended to more advanced choices, such as Metapath2Vec\cite{dong2017metapath2vec} to further improve the performance. In this work, we just use Node2Vec as a simple example. 

To build a new graph that can better facilitate downstream tasks, the topology embeddings also need to include downstream tasks' relevant information. We propose to incorporate such information by constraining the downstream task label of each node in the sampled walk to be the same. In practice, we put the nodes existing in the sampled walks in each batch and their neighbor nodes into the networks and learn topology embedding with the above objective function.

\begin{figure}[tb]
    \centering
    \includegraphics[width=1.0\linewidth]{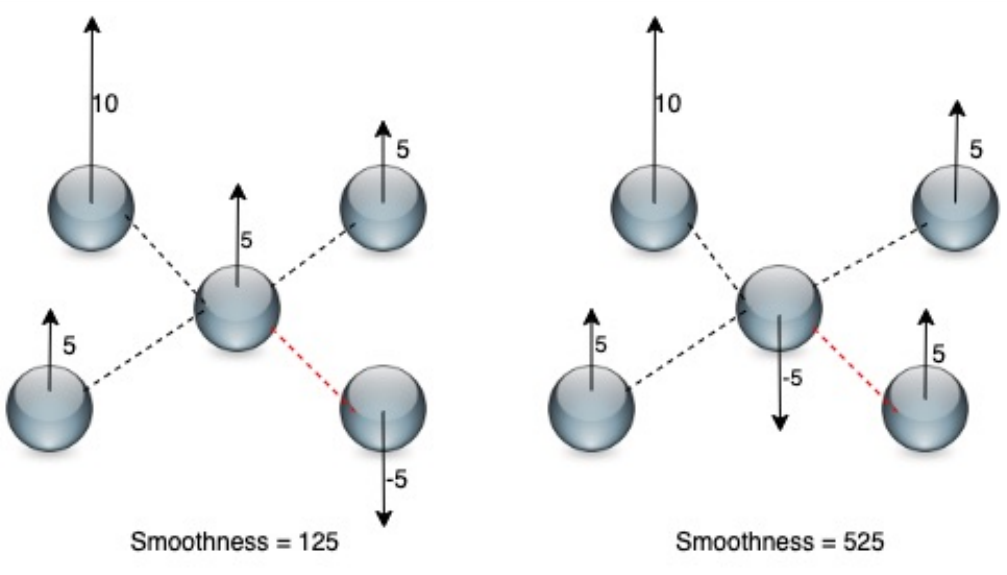}
    \caption{An illustration of smoothness on graphs. The arrows represent node signals. The two graphs has the same graph signals while having different graph structures. The left one is smoother than the right one. In ToGRL, we treat the topology embeddings as the node signals and aim to construct a smooth graph.}
    \label{fig:smoothness}
\end{figure}

Containing latent topology information and downstream tasks information, the topology embeddings are then utilized to construct a new graph. In Graph Signal Processing (GSP), a typical approach is to infer a graph with the hypothesis that the \textit{smoothness} of observation with respect to underlying graph structure \cite{dong2019learning,humbert2021learning}. In GSP, given graph signal $y\in R^{N}$, the smoothness is defined as follows:
\begin{equation}
    Smoothness = y^T \textbf{L} y = \frac{1}{2}\sum_{(i, j)\in \mathcal{E}}w_{ij}(y_i - y_j)^2
\end{equation}
where $w_{i,j}$ is the weight of edge $(i, j)$ and $\textbf{L}=\textbf{D}-\textbf{A}$ is the Laplacian matrix of graph $\mathcal{G}$. A visual illustration is given in Fig.~\ref{fig:smoothness}. 

Specifically, this smoothness requires that the distance between nodes along each edge be as small as possible. Inspired by this hypothesis, we build a new graph with a similar intuition. In this work, we treat the topology embeddings as the graph signals of the new graph to be inferred. To obtain a smooth and sparse graph, we calculate the Euclidean distance between all target nodes and preserve the top-k closest neighbors for each node. The new graph $\mathcal{\tilde{G}}=(\mathcal{\tilde{V}}, \mathcal{\tilde{E}}, \tilde{X})$ is defined as follows:
\begin{equation}
\begin{split}
       \mathcal{\tilde{V}} &= \{v| v \; \in \; \mathcal{V}_{tar}\},\;  \\
       \mathcal{\tilde{E}} &= \{(v, u)|v\; \in \; \mathcal{\tilde{V}}, \;u \; \in\; \{ Top-k\; \textrm{closest to} ||f(v)-f(u)||^2_2\} \}, \; \\
       \mathcal{\tilde{X}} &= \{f(v)| v \in \mathcal{\tilde{V}}\}
\end{split}
\end{equation}
 and the node feature matrix $\tilde{\textbf{X}}$ is the topology embedding matrix. 
\subsection{Graph Representation Learning}
In this part, a representation learning module takes input as the new graph $\mathcal{\tilde{G}}$ and learns the final node embeddings. CKD \cite{wang2022collaborative} utilizes knowledge distillation to learn node embeddings and achieves promising results. The CKD model takes input as several meta-paths induced homogeneous graphs which are first processed by graph diffusion. Knowledge distillation is performed on sub-graph level embeddings named regional knowledge, and meta-path level embeddings named as knowledge with both intra-meta-path collaborative distillation and inter-meta-path collaborative distillation. Different from CKD, our embedding just takes input as one graph, thus we only apply its intra-meta-path distillation. 

In our representation learning module, given the new graph $\mathcal{\tilde{G}}$ , the embedding of each node is firstly learned by a GCN based on its one-hop sub-graph. In each subgraph, only the centered node embedding $h_i, i\in \mathcal{V}$, is used for further computing and downstream tasks. Context sub-graph embedding is modeled as regional knowledge reflecting the relationships among the neighbors of the central node, which is defined as follows:
\begin{equation}
    l_i = \sigma\left(\frac{1}{K}\sum_{j\in \mathcal{V}_{\mathcal{\tilde{G}}_i}}h_j\right)
\end{equation}
where $\sigma$ is the sigmoid function $\sigma(x) = 1/(1+\mathrm{exp}(-x))$ and $\mathcal{V}_{G_i}$ is the neighbor node set of centered node $i$. To preserve the global connection pattern, graph embedding is modeled as the global knowledge, which is defined as follows:
\begin{equation}
    p = \sigma\left(\frac{1}{|\mathcal{\tilde{V}}|}\sum_{j\in \mathcal{\tilde{V}}}h_j\right)
\end{equation}
The training objective is a collaborative knowledge distillation over the node embedding, regional knowledge and global knowledge within the input graph $\tilde{G}$. The distillation is measured by Mutual Information (MI) and the objective function is defined as follows:
\begin{equation}
    \mathcal{L} = -\sum_{i \in \mathcal{\tilde{V}}}\left(\mathrm{MI}(h_i, l_i)+\mathrm{MI}(h_i, p)\right)
\end{equation}
Following CKD, we also apply the Jensen-Shannon divergence (JSD) to estimate the mutual information, which can be expressed as:
\begin{equation}
    MI(U,V) = \mathbb{E_P}\left[-\mathrm{sp}(-g(u,v))\right]+
    \mathbb{E_{P\times\hat{P}}}\left[\mathrm{sp}(g(u,\hat{v}))\right]
\end{equation}
where sp is the soft plus function $\mathrm{sp}(x)=\mathrm{log}(1+e^x)$, $g()$ is the representation learning model, $\mathbb{P}$ and $\mathbb{\hat{P}}$ denote the distributions of positive and negative samples, respectively. For regional knowledge, the neighbors of each centered node are regarded as positive samples and the randomly sampled other nodes are negative samples. For global knowledge, the graph embedding $p$ is the positive sample and the graph embedding learned from a graph with corrupted node features is the negative sample. 

\subsection{Prompt Decoder}
To evaluate the performance of node embeddings on the downstream tasks, traditionally a simple linear decoder such as SVM is applied to fit on the test data and give the final score. However, most graph tasks are semi-supervised tasks and the training samples for the decoder are usually inadequate and the downstream task decoder cannot be well trained.


In the HRGL task, we observe that similar to the `pre-training, fine-tuning' framework in NLP, there also exists a gap between learning node embeddings and training on downstream tasks. However, different from natural language, graph tasks are abstract, thus we cannot manually design a hard prompt. By contrast, soft prompt \cite{zhong2021factual,li2021prefix} can automatically adjust the prompt with a more flexible advantage than hard prompts and it is more suitable for graph tasks. To this end, we propose to utilize prefix soft prompt tuning \cite{zhong2021factual,li2021prefix} in downstream task adaptation of heterogeneous graphs.
A task-specific trainable vector is applied as the prefix of the original embeddings. The objective is defined as follows:
\begin{equation}
    \max_{\beta,\theta}\mathrm{log}P(y|h, \beta,\theta) = 
    \max_{\beta,\theta}\sum_i \mathrm{log}P(y_i|[\beta||h_i],\theta)
\end{equation}
where $\beta$ is the prompt embedding, $\theta$ is the parameter of a downstream task decoder and $|\cdot|$ is a concatenation operation. Then we let the decoder automatically learn a task-specific prompt to reformulate the downstream task similar to the representation learning objective together with its own parameters. In this way, we improve the performance of the decoder without changing itself.  

\section{Experiments}
In this section, we conduct extensive experiments on five real-world datasets on node classification and link prediction tasks together with detailed analysis. We aim to study the following questions:

\textbf{RQ1}. Does the ToGRL framework outperform baselines?

\textbf{RQ2}. Does the learned new graph contain enough information for downstream tasks?

\textbf{RQ3}. Does the prompt tuning decoder narrow the gap between representation learning and downstream tasks?
\subsection{Datasets}
Extensive experiments are performed on five real-world datasets to evaluate ToGRL's performance, which are ACM, ACM2, DBLP, IMDB and PubMed. Please note that we use the same version of ACM, DBLP and IMDB as those in GTN \cite{yun2019graph} and ACM2 and PubMed are the CKD's \cite{wang2022collaborative} versions. The statistics of these datasets are summarized in Table~\ref{tab:datasets_summary}.
\begin{table}[]
    \centering
    \caption{Summary of datasets statistics used in experiments.ACM, IMDB, DBLP and ACM2 are used for node classification tasks. ACM2 and PubMed are used for the link prediction task.}
    \begin{tabular}{cccccc}
    \toprule
         Datasets & Nodes & Edges & Node Types & Relations & Labels\\
         \midrule
         ACM & 8,994 & 12,961 & 3 & 4 & 3 \\
         IMDB & 12,772 & 18,644 & 3 & 4 & 3 \\
         DBLP & 18,405 & 67,946 & 3 & 4 & 4 \\
         ACM2 & 29,930 & 61,770 & 2 & 2 & 7 \\
         PubMed & 63,109 & 125,167 & 4 & 10 & 8\\
         \bottomrule
    \end{tabular}
    \label{tab:datasets_summary}
\end{table}

\begin{table*}[tp]
    \centering
    \caption{Node classification results on the four real-world datasets. Bold fonts denote the best performance among all methods. `-' denotes that the model cannot be run on the dataset on a 16GB GPU.}
    \setlength{\tabcolsep}{1.80mm}{
\begin{tabular}{c|cc|cc|cc|cc}
\hline
Dataset        & \multicolumn{2}{c|}{ACM}                                                  & \multicolumn{2}{c|}{IMDB}                                                  & \multicolumn{2}{c|}{DBLP}                                                 & \multicolumn{2}{c}{ACM2}                                                   \\ \hline
Metrics        & \multicolumn{1}{c|}{Marco F1}                 & Micro F1                  & \multicolumn{1}{c|}{Marco F1}                  & Micro F1                  & \multicolumn{1}{c|}{Marco F1}                 & Micro F1                  & \multicolumn{1}{c|}{Marco F1}                  & Micro F1                  \\ \hline
Transh         & \multicolumn{1}{c|}{$67.21 \pm 0.20$}         & $67.11 \pm 0.19$          & \multicolumn{1}{c|}{$30.78 \pm 0.23$}          & $53.61 \pm 0.20$          & \multicolumn{1}{c|}{$27.42 \pm 0.41$}         & $34.05 \pm 0.68$          & \multicolumn{1}{c|}{$17.73 \pm 0.61$}          & $50.78 \pm 0.10$          \\
Node2Vec       & \multicolumn{1}{c|}{$81.57 \pm 0.39$}         & $81.22 \pm 0.41$          & \multicolumn{1}{c|}{$48.59 \pm 1.16$}          & $64.59 \pm 0.41$          & \multicolumn{1}{c|}{$93.26 \pm 0.28$}         & $94.05 \pm 0.06$          & \multicolumn{1}{c|}{$60.01 \pm 0.61$}          & $74.61 \pm 0.46$          \\
Metapath2Vec   & \multicolumn{1}{c|}{$78.59 \pm 0.30$}         & $78.11 \pm 0.30$          & \multicolumn{1}{c|}{$54.89 \pm 0.72$}          & $67.16 \pm 0.92$          & \multicolumn{1}{c|}{$92.67 \pm 1.01$}         & $93.52 \pm 0.91$          & \multicolumn{1}{c|}{$57.37 \pm 1.11$}          & $72.64 \pm 1.32$          \\ \hline
SLAPS          & \multicolumn{1}{c|}{$88.51 \pm 0.51$}         & $88.52 \pm 0.66$          & \multicolumn{1}{c|}{$57.89 \pm 0.24$}          & $61.35 \pm 0.18$          & \multicolumn{1}{c|}{$55.32 \pm 1.20$}         & $56.70 \pm 1.11$          & \multicolumn{1}{c|}{$61.22 \pm 0.73$}          & $72.85 \pm 0.14$          \\
HGSL           & \multicolumn{1}{c|}{$93.21 \pm 0.59$}         & $93.17 \pm 0.59$          & \multicolumn{1}{c|}{$43.34 \pm 0.21$}          & $58.83 \pm 0.25$          & \multicolumn{1}{c|}{-}                        & -                         & \multicolumn{1}{c|}{-}                         & -                         \\ \hline
GTN            & \multicolumn{1}{c|}{$91.51 \pm 0.27$}         & $91.49 \pm 0.27$          & \multicolumn{1}{c|}{$53.57 \pm 2.08$}          & $55.51 \pm 1.94$          & \multicolumn{1}{c|}{$90.56 \pm 0.13$}         & $91.55 \pm 0.12$          & \multicolumn{1}{c|}{$61.34 \pm 1.14$}          & $77.92 \pm 0.12$          \\
HGT            & \multicolumn{1}{c|}{$89.56 \pm 0.51$}         & $89.47 \pm 0.55$          & \multicolumn{1}{c|}{$39.91 \pm 0.29$}          & $53.52 \pm 0.31$          & \multicolumn{1}{c|}{$85.01 \pm 0.41$}         & $86.31 \pm 0.42$          & \multicolumn{1}{c|}{$64.62 \pm 1.38$}          & $78.39 \pm 0.23$          \\
SimpleHGN      & \multicolumn{1}{c|}{$91.52 \pm 0.25$}         & $91.45 \pm 0.27$          & \multicolumn{1}{c|}{$41.97 \pm 0.66$}          & $56.12 \pm 0.89$          & \multicolumn{1}{c|}{$91.72 \pm 0.18$}         & $92.76 \pm 0.19$          & \multicolumn{1}{c|}{$68.62 \pm 0.61$}          & $79.25 \pm 0.04$          \\
CKD            & \multicolumn{1}{c|}{$90.25 \pm 0.22$}         & $90.11 \pm 0.22$          & \multicolumn{1}{c|}{$6.73 \pm 0.20$}           & $64.64 \pm 0.22$          & \multicolumn{1}{c|}{$86.54 \pm 0.05$}         & $87.89 \pm 0.10$          & \multicolumn{1}{c|}{$65.87 \pm 0.60$}          & $78.03 \pm 1.20$          \\
SeHGNN         & \multicolumn{1}{c|}{$91.48 \pm 0.28$}         & $91.54 \pm 0.28$          & \multicolumn{1}{c|}{$55.29 \pm 0.70$}          & $54.29 \pm 0.67$          & \multicolumn{1}{c|}{$93.32 \pm 0.61$}         & $92.81 \pm 0.67$          & \multicolumn{1}{c|}{$\textbf{78.00} \pm \textbf{0.18}$} & $69.98 \pm 0.17$          \\ \hline
\textbf{ToGRL} & \multicolumn{1}{c|}{$\textbf{93.38}\pm \textbf{0.11}$} & $\textbf{93.31} \pm \textbf{0.10}$ & \multicolumn{1}{c|}{$\textbf{66.88} \pm \textbf{0.25}$} & $\textbf{70.15} \pm \textbf{0.20}$ & \multicolumn{1}{c|}{$\textbf{94.90}\pm \textbf{0.12}$} & $\textbf{95.45} \pm \textbf{0.13}$ & \multicolumn{1}{c|}{$72.44 \pm 0.71$}          & $\textbf{80.67} \pm \textbf{0.89}$ \\ \hline
\end{tabular}%
}
    \label{tab:results}
\end{table*}

\textbf{ACM} is a citation network dataset which has three types of nodes, i.e., Author (A), Paper (P) and Subject (S). It also has four relations, which are PA, AP, PS and SP. The labels are the categories of papers.

\textbf{IMDB} is a movie dataset containing three node types: Movies (M), Actors (A) and Directors (D), and 4 relations: MA, AM, MD and DM. The labels are the genres of movies. 

\textbf{DBLP} is another citation network dataset that involves three node types, which are Author (A), Paper (P), Conference (C) and 4 relations, which are AP, PA, PC, CP. Different from ACM, the labels are the research areas of the authors. 

\textbf{ACM2} is also a citation network dataset containing two node types, Author (A), Paper (P) and 2 relations, which are AA, AP. The labels are the research areas of the authors. 

\textbf{PubMed} is a medical network that contains four types of nodes, including Gene (G), Disease (D), Chemical (C) and Species (S), and 10 relations. The labels are the categories of disease. 

\subsection{Baseline Methods}\label{sec:baseline}
We compare ToGRL with one typical relation learning method, one random walk based method, two graph structure learning methods, and state-of-the-art HGRL methods. 

\textbf{Transh} \cite{wang2014knowledge}: Transh is a relation learning method that models the relations as hyperplanes and utilizes the translation distance to learn node embeddings together with relation embeddings. 

\textbf{Node2Vec} \cite{grover2016node2vec}: Node2Vec is a random walk-based method that adopts a strategy to control the walking preference between DFS and BFS.

\textbf{Metapath2Vec} \cite{dong2017metapath2vec}: Metapath2Vec performs random walk based on a given meta-path and is trained with skip-gram loss.

\textbf{SLAPS \cite{fatemi2021slaps}}: SLAPS is a homogeneous graph structure learning method that jointly learns a new adjacency matrix and downstream tasks together with a node feature reconstruction task. 

\textbf{HGSL} \cite{zhao2021heterogeneous}: HGSL is a heterogeneous graph structure learning method. Feature similarity graphs, feature propagation graphs and semantic graphs are constructed and fused with the original raw graph structure by learning attention weights for different graphs.

\textbf{GTN} \cite{yun2019graph}: GTN generates multiple new meta-path adjacency matrices through its graph transformer layer and learns node embeddings from different matrices.

\textbf{HGT} \cite{hu2020heterogeneous}: HGT follows message-passing and applies Transformer~\cite{vaswani2017attention} to learn the attention weights among neighbors. 

\textbf{SimpleHGN} \cite{lv2021we}: SimpleHGN extends GAT by introducing relation embeddings into the process of calculating attention weights and achieving promising results recently.

\textbf{CKD} \cite{wang2022collaborative}: CKD utilizes knowledge distillation and simultaneously extracts information from different meta-path adjacency matrices.

\textbf{SeHGNN} \cite{yang2023simple}: SeHGNN is the SOTA model which is designed in a single-layer structure with long metapaths. To reduce complexity, it pre-computes the neighbor aggregation by a mean aggregator and removes overused neighbor attention. 

\subsection{Experimental Settings}
 We use the hyperparameters that achieve the best performance on each dataset for baseline models. We also choose popular meta-paths as used in previous works \cite{wang2022collaborative,zhao2021heterogeneous,yun2019graph} for meta-path-based models. In terms of ToGRL, we apply GAT as the backbone for ACM, DBLP and IMDB datasets and GCN for ACM2 and PubMed datasets for topology embedding learning. We set the learning rate in \{0.001, 0.0001, 0.00001\} for both the graph structure learning module and the representation learning module. The value $k$ is searched from \{15, 20, 25, 30, 35, 40\}. A linear classifier is selected as the downstream task decoder. The whole model is implemented with PyTorch.
 
\subsection{Node Classification}
In this section, we select node classification as the downstream task and evaluate ToGRL performance. We follow the experimental settings as used in CKD \cite{wang2022collaborative}. The test dataset is randomly split into 5 folds for cross-validation for unsupervised models.
For supervised models, i.e. SLAPS, HGSL, GTN, HGT, SImpleHGN and SeHGNN, we train them using the train sets and validation sets in the normal way. Then we test them in the same way as unsupervised models using the learned node embeddings. 
The downstream task decoder is trained on one fold and tested on the other folds. We select Macro F1 and Micro F1 as evaluation metrics and report the average scores. The details of the results are summarized in Table~\ref{tab:results}. Based on the experimental results, we have three major findings as follows:

(1) Compared with baseline models, our ToGRL can generally achieve the best performance on all four datasets. Specifically, ToGRL outperforms the existing best baselines by 6.88\% on Marco F1 and 5.51\% on Micro F1 on the IMDB dataset. The consistent improvements on all datasets convincingly demonstrate the effectiveness of this proposed learning framework with a novel graph structure learning module and prompt tuning decoder. 

(2) Compared with CKD, our model only takes input as one graph while CKD uses multiple meta-paths induced homogeneous graphs. The improvement shows that our new input graph is superior to the original meta-paths graphs which can better facilitate downstream tasks and contains necessary information related to each meta-path. 

(3) Among four datasets, the improvements made by ToGRL on IMDB and ACM2 are the most obvious compared with none graph structure learning methods. The observation that the average scores on these datasets (i.e., IMDB, ACM2) are lower than those of the other two datasets (i.e., ACM, DBLP) indicates that these two datasets have more noise. Thus, the significant improvements made on these two datasets show the necessity of learning a new graph. Furthermore, compared with graph structure learning baselines (i.e., HGSL, SLAPS), the improvement shows that our ToGRL can better extract latent relationship between nodes and it is trained in a simpler and more stable way. Specifically, the failure of HGSL on DBLP and ACM2 datasets shows that HGSL suffers from memory consumption problem. It also performs poorly on IMDB dataset. This may be attributed to the fact that HGSL relies on fixed feature similarity graph and  feature propagation graphs, while the node features of IMDB are noisy discrete vectors leading to possible corruption of those feature graphs. Besides, the usage of a semantic graph based on pre-trained Metapath2Vec embeddings also limits its performance.  

(4) We observe that random walk based methods (i.e., Node2Vec, Metapath2Vec) perform well on DBLP dataset. This indicates that the classification task on DBLP dataset relies more on long-term dependencies and these two methods can better capture this relation. Besides, we observe that the SOAT baseline method SeHGNN obtains a high score on Macro F1 while achieving a relatively low score on Micro F1 on the ACM2 dataset, which indicates it suffers imbalanced learning on different categories.

\subsection{Ablation Study}
In order to verify that the design of each part of ToGRL is effective, we design several variants of ToGRL. We compare these variants with the best score selected from baselines. The results are summarized in Table~\ref{tab:ablation}.

\begin{itemize}
    \item \textbf{ToGRL-p\&w/o prompt} drops prompt tuning and utilizes inner product when constructing the new graph. 
    \item \textbf{ToGRL-p} applies inner-product distance when constructing the new graph. 
    \item \textbf{ToGRL-w/o prompt} applies only linear downstream task decoder without the prompt tuning technique.
\end{itemize}
By comparing between ToGRL-w/o prompt with the original ToGRL, and ToGRL-p with ToGRL-p\&w/o prompt, we find that our proposed prompt tuning is effective and narrows the gap between embedding learning and downstream tasks. The decline of  ToGRL-p with ToGRL-p\&w/o prompt compared with ToGRL-w/o prompt shows that the choice of using Euclidean distance is slightly better than the inner-product distance. We point out that in some cases, although the value of the inner product distance between two embeddings is large, the Euclidean distance of these two can be large. Besides, the Euclidean distance is much more similar to the original smoothness definition than the inner-product distance, thus it is a better choice.  
\begin{table}[tp]
    \centering
    \caption{The ablation study results. Bold fonts represent the best performance among ToGRL variants and baselines.}
    \begin{adjustbox}{width=0.48\textwidth}
    \begin{tabular}{c|cccc}
    \toprule
                  & \multicolumn{2}{c}{IMDB} & \multicolumn{2}{c}{DBLP} \\
                 & Macro F1    & Micro F1   & Macro F1    & Micro F1   \\
    \midrule
Baseline & 60.92       & 62.59      & 93.26       & 94.05      \\
ToGRL-p\&w/o prompt  & 61.85       & 67.93      & 93.89       &94.60  \\
ToGRL-product    & 65.02           & 67.85      & \textbf{94.89}           & \textbf{95.48 }         \\
ToGRL-w/o prompt & 64.25       & 69.64      & 94.24       & 94.96      \\
ToGRL            & \textbf{66.88}       & \textbf{70.15}      & \textbf{94.90}       & \textbf{95.45}      \\
\bottomrule
\end{tabular}
\end{adjustbox}
    \label{tab:ablation}
\end{table}

\begin{table}[tp]
\caption{The AUC scores on the Link Prediction task. Bold fonts denote the best performance among all methods.`-' denotes that the model cannot be run on the dataset on a single 16GB GPU card. }
\begin{tabular}{c|cc}
\toprule
             & ACM2  & PubMed \\ \midrule
Metapath2Vec & 0.712 & 0.628  \\
SLAPS        & 0.662    &    -    \\
HAN          & 0.868 & 0.717  \\
HGT          & 0.920 & 0.736  \\
CKD          & 0.948 & 0.735  \\
\textbf{ToGRL}        & \textbf{0.951} & \textbf{0.796} \\ \bottomrule
\end{tabular}
\label{tab: lp}
\end{table}

\begin{figure}[t]
  \centering
  \includegraphics[width=1.0\linewidth]{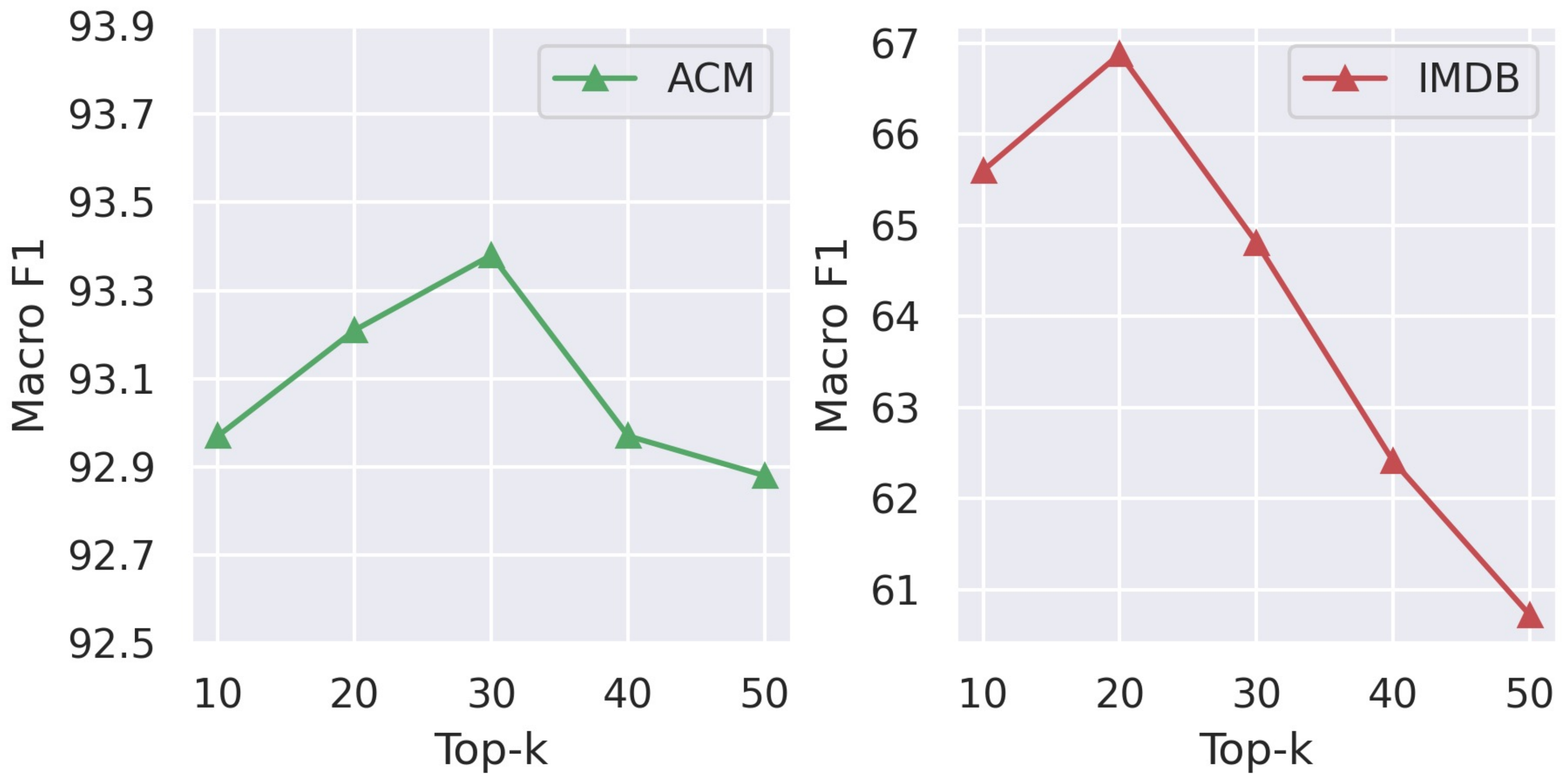}
  \caption{The performance of ToGRL on ACM and IMDB datasets with different $k$ values.}
  \label{fig:para_analysis}
\end{figure}

\begin{table}[tp]
    \caption{Memory consumption results (GB) for different models on IMDB and ACM datasets on node classification. The ToGRL\_st represents the structure learning part in our framework and the ToGRL\_re represents the representation learning part. The model settings are the same as those reported in Table \ref{tab:results}.}
    \setlength{\tabcolsep}{1.0mm}{
    \begin{tabular}{c|cc|cc|cc}
\hline
     & SLAPS & HGSL & GTN & SimpleHGN & ToGRL\_st & ToGRL\_re \\ \hline
IMDB & 9.4   & 11.7 & 2.7 & 2.1       & 2.7       & 1.7       \\
ACM  & 5.2   & 8.6  & 1.6 & 2.0       & 2.1       & 1.6       \\ \hline
\end{tabular}%
}
\label{tab: memory}
\end{table}

 \begin{figure}[t]
    \centering
    \includegraphics[width=1.0\linewidth]{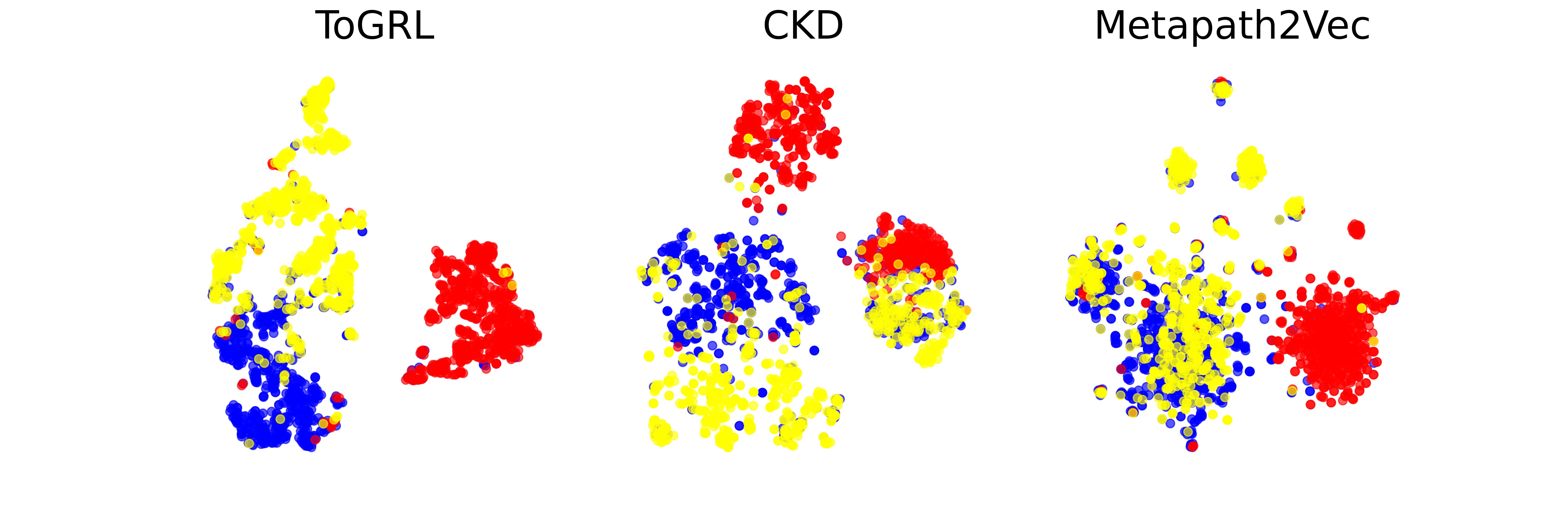}
    \caption{Visualization of node embeddings on ACM dataset. Compared with baseline models, our ToGRL can successfully learn more separated node embeddings which further proves that our structure learning module can  extract downstream tasks related topology information effectively and facilitate the downstream task eventually. }  
    \label{fig:my_label}
\end{figure}

\subsection{Link Prediction}

We further evaluate our model on the link prediction task, which is another basic downstream task usually used to evaluate node embeddings. In our experiments, we follow the experiment settings used in the CKD \cite{wang2022collaborative}. Given a raw heterogeneous graph, the training graph is generated by first selecting a certain type of relation and removing 20\% of edges belonging to this type as missing edges. Then the model is trained on the training graph to obtain node representations. To evaluate the models' performance on link prediction, we predict node pairs that tend to have missing edges and report AUC scores. In this work, we select typical baseline models of each category mentioned in the section \ref{sec:baseline} to perform the link prediction task on two real-world datasets. The results are shown in Table \ref{tab: lp}. From the table, we have the following observations:

\begin{itemize}
\item  Among the baseline models, our ToGRL still performs the best on the two real-world link prediction datasets. This further proves the effectiveness of our framework. 
\item   Similar to the performance on node classification, the previous graph structure learning baseline method, i.e. SLAPS performs poorly on the ACM2 dataset, which further reveals that our ToGRL can better extract useful topology information. 



\end{itemize}

\subsection{Parameter Analysis}
In this section, we study the impacts of the hyperparameter $k$ used to construct graphs in the proposed framework. We report Macro F1 score of ToGRL on ACM and IMDB with different $k$ values varying from 10 to 50. The results are shown in Fig.~\ref{fig:para_analysis}.  In Fig.~\ref{fig:para_analysis}, we observe that the performance of ToGRL reaches the top when $k$ is around 30. This indicates that ACM has less noise. For IMDB, the performance first increases and reaches the peak at $k=20$, then it drops sharply. A possible explanation is that there exists more noise in the IMDB, and with $k$ increasing, redundant information is introduced into each node.

\subsection{Memory Consumption}
 Based on the models' performance on the node classification and link prediction task, we find that our ToGRL can also achieve good performance in terms of memory consumption while both the graph structure learning baseline models suffer severe memory issues. To further verify this conclusion, we compare our ToGRL with baseline models on memory consumption issues and details are given in Table \ref{tab: memory}.
 Specifically, compared with graph structure learning baselines, i.e.SLAPS and HGSL, our framework consumes much less memory in both the two learning stages. Besides, our framework also achieves comparable performance compared with normal HGSL methods, i.e. GTN and Simple HGN, which directly learn node embeddings on the raw graphs. In summary, our framework can achieve better results on downstream task and preserve low memory consumption in the meantime, which further proves the efficiency of our framework.

\subsection{Embeddings Visualization}
This section provides a visualization of the learned embeddings on the ACM dataset. We use t-SNE \cite{van2008visualizing} to project the embeddings into a 2-D space and plot the results in Fig.~\ref{fig:my_label}. The results show that our ToGRL performs well in classifying the nodes with different classes. Specifically, compared with CKD, our ToGRL uses a similar representation learning method but incorporates a new graph structure learning module. The improvement further proves that ToGRL successfully learns a new graph containing the downstream task information.

\section{Conclusion}
In this paper, we have designed a novel heterogeneous graph representation learning method, named ToGRL, which proposes a novel graph structure learning module to automatically mine latent graph topology information together with downstream tasks. The newly learned graph contains rich information and can better facilitate downstream tasks. This novel graph structure learning module is also memory efficient. Moreover, we propose to leverage a prompt tuning method to close the gap between representation learning and downstream task learning. We perform extensive experiments on five real-world datasets on node classification and link prediction tasks to demonstrate the effectiveness of the proposed ToGRL method.

\bibliographystyle{ACM-Reference-Format}
\bibliography{sample-base}

\appendix









\end{document}